\documentclass{article}

\PassOptionsToPackage{numbers}{natbib}

\usepackage[main, final]{neurips_2026}

\usepackage[utf8]{inputenc} 
\usepackage[T1]{fontenc}    
\usepackage{hyperref}       
\usepackage{url}            
\usepackage{booktabs}       
\usepackage{amsfonts}       
\usepackage{nicefrac}       
\usepackage{microtype}      
\usepackage[table]{xcolor}

\usepackage{caption} 
\usepackage{amsmath}
\usepackage{amssymb}
\usepackage{amsfonts}
\usepackage{graphicx}
\usepackage{caption}
\usepackage{subcaption}
\usepackage{algorithm}
\usepackage{algpseudocode}
\usepackage{wrapfig}
\usepackage{caption}

\definecolor{rowhighlight}{RGB}{234,240,249} 

\title{Probabilistic Tiny Recursive Model}

\author{
  Amin~Sghaier \\
  Mila -- Quebec AI Institute\\
  ILLS \& ETS Montreal \\
  \And
  Ali~Parviz \\
  Mila -- Quebec AI Institute  \\
  UC San Diego 
  \And
  Alexia~Jolicoeur-Martineau \\
  Independent \\
  \AND
  \texttt{\{amin.sghaier, ali.parviz\}@mila.quebec} \\
  \texttt{alexia.jolicoeur-martineau@mail.mcgill.ca}
}

\begin{document}


\maketitle

\begin{abstract}

Tiny Recursive Models (TRM) solve complex reasoning tasks with a fraction of the parameters of modern large language models (LLMs) by iteratively refining a latent state and final answer. While powerful, their deterministic recursion can lead to convergence at suboptimal solutions, without escape mechanism. A common workaround relies on task-specific input perturbations at test time combined with answer aggregation via voting. We introduce \textbf{Probabilistic TRM (PTRM)}, a task-agnostic framework for test-time compute scaling that addresses this limitation through stochastic exploration. PTRM injects Gaussian noise at each deep recursion step, enabling parallel trajectories to explore diverse solution basins, and selects among them with a criterion such as the model’s existing Q head (used for early stopping in the original TRM). Without requiring retraining or task-specific augmentations, PTRM enables substantial accuracy gains across benchmarks, including Sudoku-Extreme ($87.28\%$ to $98.75\%$),  N-Queens ($48.1\%$ to $75.5\%$ on 16×16 boards), and various puzzles from Pencil Puzzle Bench ($62.6\%$ to $91.2\%$). On the latter, PTRM achieves nearly double the accuracy of an ensemble of frontier LLMs ($91.2\%$ vs. $55.1\%$) at a small fraction of the cost, using only 7M parameters. Code is available at \url{https://github.com/amins01/ptrm}.
\end{abstract}

\vspace{1em}
\begin{center}
    \includegraphics[width=\linewidth]{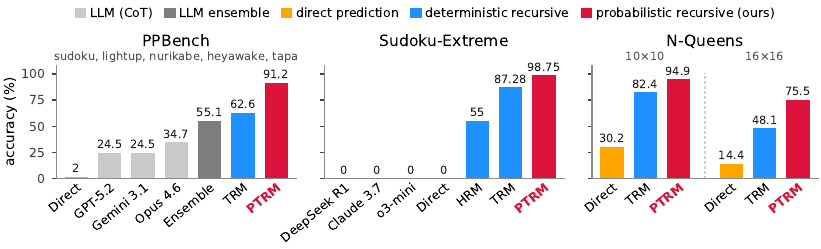}
    \captionof{figure}{\textbf{PTRM performance comparison.} On various PPBench puzzles, PTRM ($K{=}100, D{=}48$) boosts TRM ($D{=}16$) performance by $28.6$ points without any retraining. It outperforms the strongest single frontier LLMs by $56.5$ points and an ensemble of the seven strongest LLMs (assuming a perfect verifier) by $36$ points. On Sudoku-Extreme, PTRM reaches a state of the art $98.75\%$, and on N-Queens, where puzzles have many valid answers, it raises accuracy from $82.4\%$ to $94.9\%$ ($10{\times}10$) and from $48.1\%$ to $75.5\%$ ($16{\times}16$).}
    \label{fig:headline_bar}
\end{center}


\section{Introduction}\label{sec:intro}
Tiny Recursive Models (TRM)~\citep{jolicoeur2025less} achieve strong performance on complex reasoning puzzles with orders of magnitude fewer parameters than the large language models (LLMs) they outperform on tasks like Sudoku-Extreme~\citep{wang2025hierarchical} and ARC-AGI~\citep{chollet2019measure,chollet2025arc}. TRM and its predecessor Hierarchical Reasoning Model (HRM)~\citep{wang2025hierarchical} represent an emerging architectural alternative to standard autoregressive reasoning models. Rather than autoregressively generating chains of token-level reasoning, they recursively refine a latent state. This approach produces a single deterministic answer per input, fitting well with tasks where the answer is unique.

Despite their strong performance, their deterministic inference does not make full use of their capabilities. We show that many of TRM's incorrect answers are from rollouts trapped in bad latent space basins (i.e., regions of the latent space which decode to incorrect answers and from which the deterministic recursions cannot escape). This observation, which aligns with recent mechanistic work on related models~\citep{ren2026your}, suggests that TRM has the capabilities to solve significantly more problems but is limited by its standard inference procedure.

Although each puzzle has a unique correct answer, many distinct latent trajectories can reach it. This is analogous to reasoning LLMs, where many reasoning trajectories can lead to the same unique answer. However, being non-deterministic, LLMs can be randomly sampled in order to form different trajectories (including Chains of Thought and actual answer). By then selecting a trajectory using a voting mechanism or based on the answer's projected value (via a verifier), LLMs can leverage test-time compute to achieve very high accuracy~\citep{snell2024scaling}. We propose a way to achieve similar test-time scaling performance gains by sampling stochastic latent trajectories, each producing a deterministic decoded answer, and selecting among the answers using the model's own Q head.

TRM's Q head is trained jointly (as a correctness classifier) with the rest of the network and is conventionally used only at training time for Adaptive Computation Time (ACT)~\citep{graves2016adaptive}. It carries valuable information that the standard inference procedure discards.

We propose \textbf{Probabilistic TRM (PTRM)}, a test-time compute scaling framework that introduces a new width scaling axis. At inference we run $K$ parallel rollouts per puzzle, each receiving Gaussian noise injected into the latent at every deep recursion step. The noise causes rollouts to follow different latent trajectories and settle in different basins. Among the resulting candidate answers, the Q head is used to select the one most likely to be correct. \textbf{PTRM requires no training changes and no task-specific test-time augmentation}, yet, as illustrated in Figure \ref{fig:headline_bar}, delivers substantial accuracy gains across diverse reasoning benchmarks.  On puzzles with several valid answers, PTRM also finds several distinct solutions, from which the Q head is able to accurately select. The Q head can also be used to stop sampling early to make inference more efficient.

\section{Background: Tiny Recursive Model}
Tiny Recursive Model (TRM) is a single network that iteratively refines a predicted answer $y$ to a question $x$ through recursive updates of a reasoning latent $z$. Specifically, a single \textbf{latent recursion} consists of $n$ updates to the latent state $z$ followed by one update to the predicted answer $y$, all using the same two-layer network $f_{\theta}$: $z \leftarrow f_\theta(x + y + z)$ ~$n$ times, then $y \leftarrow f_\theta(y + z)$.

$f_{\theta}$ distinguishes the two update types by whether the input includes $x$. A \textbf{deep recursion} runs $T$ latent recursions in sequence, with only the final one retaining gradients, allowing the model to leverage a large effective depth while keeping training efficient.

Rather than doing one optimization step per sample, TRM is trained via deep supervision, which consists in keeping the previous latent state $z$ and answer $y$ as initialization (after being detached from the computational graph) for the next supervision step. This is done for up to $N_{sup}$ supervision steps. The loss at each step is calculated using cross entropy between the predicted answer logits $f_O(y)$ (where $f_O$ is a linear output head) and the ground truth $y_{true}$. This trains the network to progressively refine its prediction across reasoning steps. At inference, the recurrence can be unrolled for more steps than during training, providing a depth axis for test-time compute scaling (additional steps may correct otherwise-incorrect answers).

Without halting mechanism during training, each puzzle stays in the mini-batch for $N_\text{sup}$ supervision steps rather than being replaced after each one. To avoid wasting compute on already-solved samples, an Adaptive Computation Time (ACT) halting mechanism is used. This is done by adding a binary cross entropy loss between a halting logit $\hat{q}=f_Q(y)$ (where $f_Q$ is a linear Q head) and the binary exact correctness of the predicted answer $\hat{y}=\arg\max f_O(y)$: $\mathcal{L}_\text{step} = \text{CE}(f_O(y), y_\text{true}) + \text{BCE}(\hat{q}, \mathbf{1}[\hat{y} = y_\text{true}])$. The Q head thus allows the supervision loop to halt early on samples where $\mathrm{sigmoid}(\hat{q}) > 0.5$, improving data efficiency. During inference, the Q head is not used, and the model performs $N_{sup}$ supervision steps to maximize answer correctness.

While TRM is powerful, it sometimes gets stuck into incorrect solutions. In the next section, we will investigate such failure cases in order to determine a way to remedy them.

\section{Problem: When Does TRM Fail?}
\label{sec:basins}

\begin{figure}[!t]
    \centering
    \includegraphics[width=\linewidth]{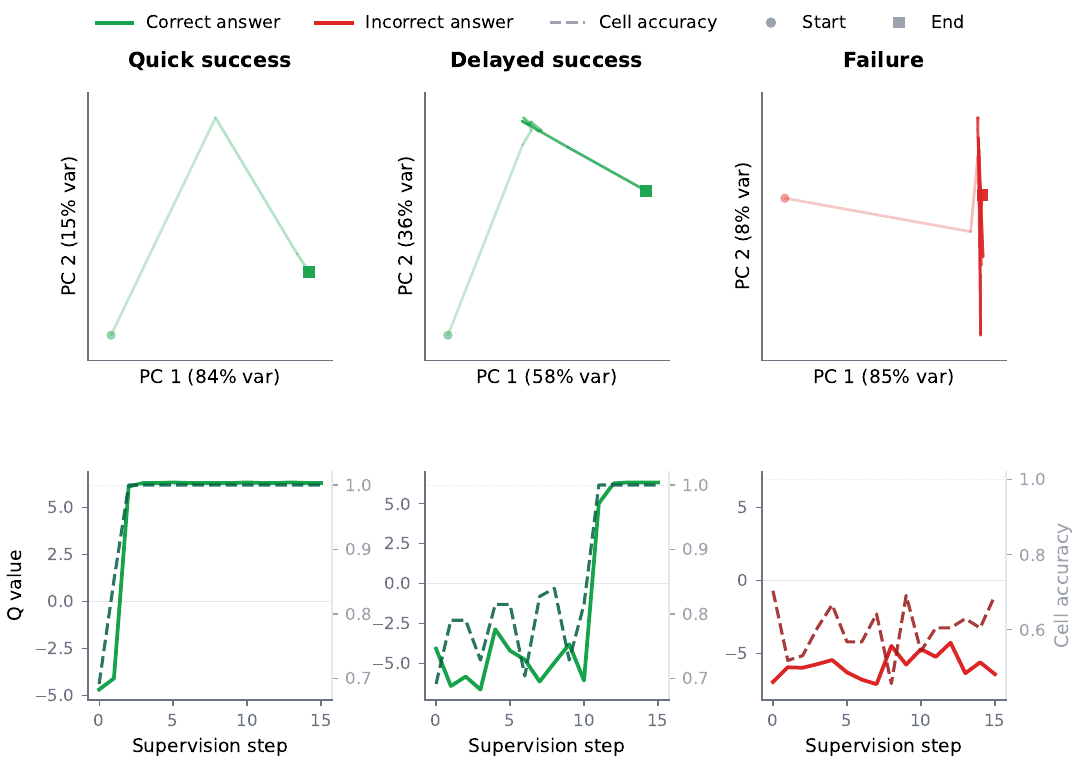}
    \caption{\textbf{TRM Trajectory Modes.}
    PCA projection of $y$ (top) and Q value (solid, left axis) with cell accuracy (dashed, right axis) across supervision steps (bottom) for three PPBench puzzles, illustrating three trajectory modes (left to right): quick success, delayed success, and failure (Sec. \ref{sec:basins}).
    Latents are projected into the principal plane per puzzle, so PC axes
    are not comparable across plots. Trajectories fade from light (early steps) to dark (later steps). Circle marks the start and square
    marks  end.
    }
    \label{fig:trajectory_modes}
\end{figure}

\subsection{Analysis of failures and successes}

We present observations about TRM that motivate our method. In this section, we train a TRM on multiple Pencil Puzzle Bench (PPBench)~\citep{waugh2026pencil} puzzles and inspect the latent dynamics and Q head behavior across supervision steps on a held-out validation set. For each puzzle, we record the latent $y_t$ and the Q logit $\hat{q}_t = f_Q(y_t)$ at every supervision step $t = 1, \ldots, N_\text{sup}$, project the latents into the principal plane (PCA per puzzle), and jointly plot the Q value alongside cell accuracy (fraction of correct cells in the predicted answer) over supervision steps. Figure~\ref{fig:trajectory_modes} shows paired PCA and Q/cell-accuracy plots for three representative puzzles, illustrating three trajectory modes we observe:

\textbf{Quick success}: the trajectory transitions in a few steps from its starting location to a convergence region and remains there. Cell accuracy and the Q value rise together and saturate near their maxima within the same few steps.

\textbf{Delayed success}: the trajectory initially oscillates around one region and remains there for multiple supervision steps before sharply escaping to a different region where it converges. During the initial phase, the Q value is negative, and at the step where the trajectory escapes, both Q value and cell accuracy spike together.

\textbf{Failure}: the trajectory oscillates in a bounded region without converging. Cell accuracy never reaches near 100\%, and the Q value stays negative for all supervision steps.

We refer to latent space regions that trajectories remain in across multiple supervision steps and exhibit similar cell accuracy throughout as basins. Basins where cell accuracy is near-maximal are good basins and basins where it is not are bad basins. Initially, failures and delayed successes behave similarly (both are caught in bad basins with negative Q). They diverge only later in their trajectories, when delayed successes find an escape to a good basin while failures remain stuck.

\subsection{The Q head tracks trajectory quality}


\vspace{4pt}
\noindent
\begin{minipage}[c]{0.62\linewidth}
    \centering
    \includegraphics[width=\linewidth]{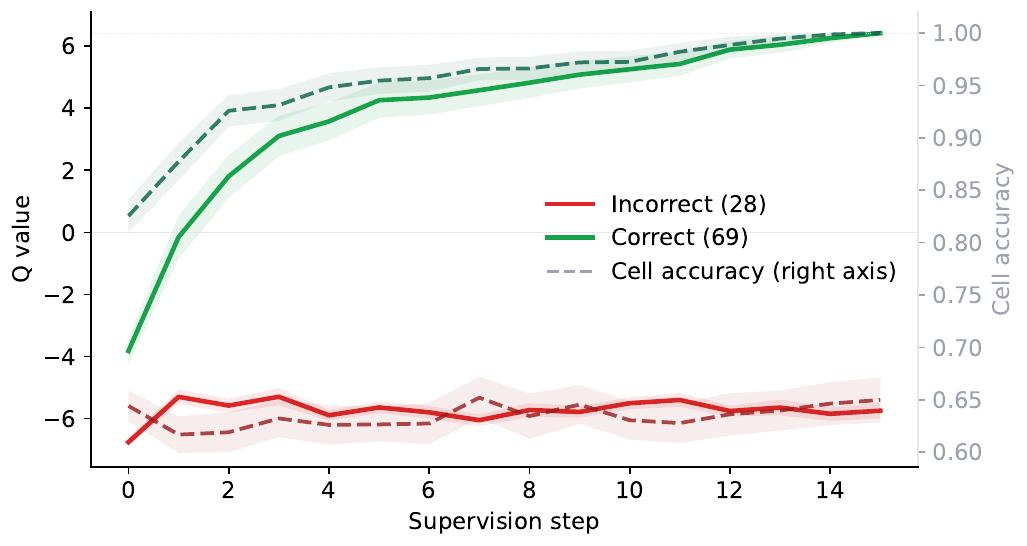}
\end{minipage}%
\hfill
\begin{minipage}[c]{0.35\linewidth}
    \captionof{figure}{\textbf{Q value follows cell accuracy across
    reasoning.} Mean Q value (solid, left axis) and mean cell accuracy
    (dashed, right axis) over supervision steps, aggregated over 100
    PPBench validation puzzles, separated by final correctness
    (green: correct, red: incorrect).}
    \label{fig:q_cellacc_aggregate}
\end{minipage}
\vspace{4pt}
Across all three modes (failures, delayed successes, and quick successes), we find that the Q head's value closely tracks cell accuracy at every supervision step. To further confirm this, Figure \ref{fig:q_cellacc_aggregate} aggregates trajectories from 100 PPBench validation puzzles, separating them by final-answer correctness. The aggregate view corroborates the per-puzzle observation: mean Q and mean cell accuracy rise together on correct trajectories and remain mostly flat on incorrect ones. Moreover, at convergence, the Q logit sharply separates the two populations where $\hat{q} \approx +6\ (\mathrm{sigmoid} \approx 1)$ for correct trajectories and $\hat{q} \approx -6\ (\mathrm{sigmoid} \approx 0)$ for incorrect ones. The Q head is therefore a reliable learned indicator of whether a trajectory has reached a good basin.

Given the Q head's ability to distinguish good from bad trajectories, a natural question follows: \emph{can we leverage the Q head to identify better trajectories?} The main challenge is that the standard TRM is inherently deterministic, and thus cannot be used to sample different trajectories for a given problem. In the next section, we will show that by simply adding Gaussian noise to the latent state, we can sample different parallel trajectories and leverage the Q head to pick the best one.

\section{Method: Test-Time Compute Scaling via Stochastic Rollouts}
\label{sec:width_scaling}

\begin{figure}[t]
\centering
\begin{minipage}[t]{0.45\linewidth}
\vspace{0pt}
\small
\hrule
\vspace{2pt}
\textbf{PTRM Inference}
\vspace{2pt}
\hrule
\vspace{4pt}

\begin{algorithmic}[1]
\State \textbf{Input:} puzzle $x$, rollouts $K$, \\ supervision steps $D$, noise scale $\sigma$, \\ selection criterion $S$
\For{$k = 1, \ldots, K$ in parallel}
    \State Initialize $z_0^{(k)}, y_0^{(k)}$
    \For{$t = 1, \ldots, D$}
        \State $z_{t-1}^{(k)} \mathrel{+}= \epsilon$, $\epsilon \sim \mathcal{N}(0, \sigma^2 I)$
        \State $z_t^{(k)}, y_t^{(k)} \gets \texttt{rec}(x, z_{t-1}^{(k)}, y_{t-1}^{(k)})$
    \EndFor
    \State $\hat{y}^{(k)} \gets \arg\max f_O(y_{D}^{(k)})$
    \State $\hat{q}^{(k)} \gets f_Q(y_{D}^{(k)})$
\EndFor
\State \Return $S\big((\hat{y}^{(k)}, \hat{q}^{(k)})_{k=1}^{K}\big)$
\end{algorithmic}

\vspace{4pt}
\hrule
\end{minipage}
\hfill
\begin{minipage}[t]{0.54\linewidth}
\vspace{0pt}
\centering
\includegraphics[width=\linewidth]{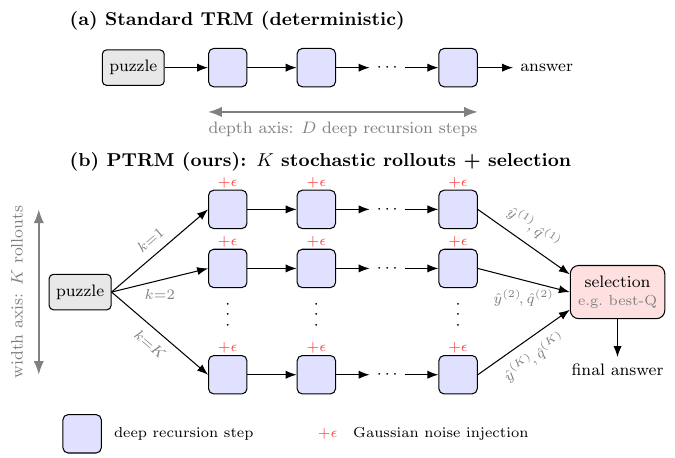}
\end{minipage}

\caption{\textbf{Left:} PTRM inference procedure (the \textit{rec()} function refers to a deep recursion step). \textbf{Right:} PTRM mechanism. (a) Standard TRM: a single deterministic rollout. (b) PTRM: $K$ stochastic latent rollouts with Gaussian noise $\epsilon$ at each deep recursion step, then select the final answer based on a criterion (we use Q head selection by default).}
\label{fig:ptrm_combined}
\end{figure}

We propose \textbf{Probabilistic TRM (PTRM)}, an inference-time procedure that makes the TRM recursion stochastic and selects the best of $K$ resulting trajectories. PTRM requires no special training and can be readily applied to any pretrained TRM model. Furthermore it requires no task-specific augmentations. PTRM works as follows: at each supervision step, we add Gaussian noise (scaled by $\sigma$) to the latent state input. The Q head $f_Q$ scores each candidate latent output, and the one with the highest Q value is selected and then decoded using the model's output head $f_O$. While we mainly focus on Q head selection, other selection criteria can be used (Appendix~\ref{app:noise-ablation}).
The algorithm in Figure \ref{fig:ptrm_combined} (left) states this formally. PTRM offers two complementary benefits: 1) it enables trajectories to escape bad basins where deterministic TRM remains stuck, and 2) it introduces \emph{width} as a new axis for test-time scaling.

\subsection{Escaping bad basins}
In Sec.~\ref{sec:basins}, we found that some failed deterministic trajectories are caught in bad solution basins in latent space, with no way to escape. PTRM lets us test whether stochastic perturbations are enough for some of the rollouts of a previously failed puzzle to reach a good solution basin.
Figure~\ref{fig:multi_rollout} shows K=100 independent rollouts, from the same failed puzzle used in Figure~\ref{fig:trajectory_modes} (which fails at K=1), projected into the principal plane. Most rollouts (92\%) remain stuck in the same bad basin, while a minority (8\%) escape to a distinct region in latent space and produce correct answers. We also observe that recurrent noise creates a per-rollout probability of escape: at $K=5$ no rollouts escape, at $K=25$ one does, and at $K=100$ eight do. This confirms that noise provides the stochasticity needed to occasionally find an escape trajectory.

\subsection{Width scaling}
Since more rollouts per puzzle compound the chance that at least one reaches a good basin, the number of rollouts $K$ is a natural quantity to scale. Given $K$ independent rollouts, pass@K (any rollout correct) is the oracle upper bound and best-Q@K (the rollout with highest $\hat{q}$ is correct) is a metric available at inference without a correctness oracle. The choice of Q as selector is motivated by Sec. \ref{sec:basins}'s observation that Q accurately separates correct from incorrect trajectories (Figure~\ref{fig:q_cellacc_aggregate}).

\begin{figure}[!t]
\centering
\begin{minipage}[t]{0.42\linewidth}
    \centering
    \includegraphics[width=\linewidth]{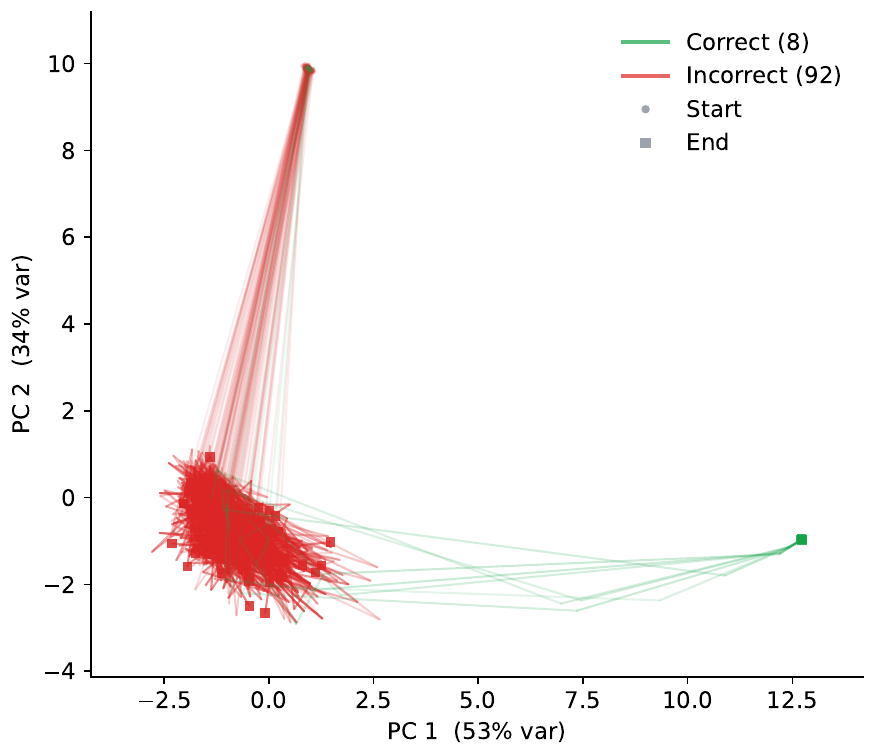}
    \captionof{figure}{\textbf{Stochastic rollouts escape bad basins.}
        Principal plane projection of $K=100$ independent rollouts of the same failed puzzle as in Figure~\ref{fig:trajectory_modes} (right). 92 rollouts remain caught in the bad basin (red). 8 escape to a good basin and produce correct answers (green).}
    \label{fig:multi_rollout}
\end{minipage}\hfill
\begin{minipage}[t]{0.48\linewidth}
    \centering
    \includegraphics[width=\linewidth]{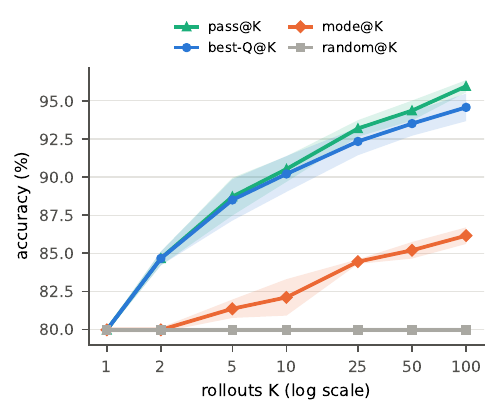}
    \captionof{figure}{\textbf{Width scaling.}
    pass@$K$, best-Q@$K$, mode@$K$, and random@$K$ as $K$ grows, averaged over 3 seeds on a held-out PPBench validation set. The Q head is a strong verifier on the tested puzzles, consistently outperforming selection of the most frequent answer.}
    \label{fig:width_scaling}
\end{minipage}
\end{figure}

Figure~\ref{fig:width_scaling} shows pass@K and best-Q@K as $K$ grows, averaged over 3 seeds on the held-out PPBench validation set (sudoku, nurikabe, tapa, lightup, and heyawake). best-Q@K rises from $80.0\%$ at $K{=}1$ to $94.6\%$ at $K{=}100$, a gain of 14.6 percentage points, while pass@K reaches $96.0\%$. Across all tested K, the gap between pass@K and best-Q@K is at most 1.4pp, making the Q head a strong verifier on this validation set. By contrast, mode@K (most frequent answer across rollouts) rises by only 6.2pp (to $86.2\%$) over the same range, and random selection stays at $80.0\%$, showing that the width-scaling gains come mostly from the Q head's ability to identify correct solutions even when they are rare.

\textbf{Interaction with depth scaling.} Depth is another scaling axis already supported by TRM, which consists of running more deep recursions (supervision steps) at inference than the $N_\text{sup}$ the model was trained on. On the deterministic baseline ($K$=1), increasing the depth from $D{=}16$ to $D{=}1600$ (matched compute with $K{=}100$ rollouts at $D{=}16$) only raises PPBench validation accuracy from $81.2\%$ to $83.4\%$, against $94.6\%$ for width scaling. Depth helps more on some puzzles like $16{\times}16$ N-Queens, but width still matches or beats it at equal compute (Sec.~\ref{sec:nqueens}). Both depth and width scaling can be seen as ways to explore the model's solution space. Since rollouts are independent and parallelizable while extra depth is sequential, width is the more practical scaling axis (Appendix~\ref{app:inference-cost}).

PTRM unlocks a simple and task-agnostic recipe for scaling TRM test-time compute. The next section evaluates the method across multiple benchmarks and against several baselines, including frontier LLMs.

\section{Experiments}
\label{sec:experiments}

This section evaluates PTRM's performance on diverse reasoning benchmarks. We compare against the deterministic TRM baseline, a non-recursive direct-prediction baseline, and frontier LLMs. Across several PPBench puzzles~\citep{waugh2026pencil}, Sudoku-Extreme~\citep{wang2025hierarchical}, Maze-Hard~\citep{wang2025hierarchical}, and ARC-AGI 2~\citep{chollet2025arc}, PTRM substantially boosts the performance of each pretrained TRM using only inference compute.

\subsection{Setup}
\label{sec:setup}

\textbf{Datasets.} Pencil Puzzle Bench (PPBench)~\citep{waugh2026pencil} consists of 62,231 constraint-satisfaction pencil puzzles (from 94 puzzle types). From the full PPBench dataset, 300 puzzles (15 puzzles from 20 types) selected by Waugh \citep{waugh2026pencil} are held out to form the golden set. From the remainder we hold out a fixed-size validation set of 100 puzzles per puzzle type (50 for tapa, due to its smaller base size), and the rest forms the training set. We filter all three sets to puzzles of six types (sudoku, lightup, nurikabe, shakashaka, heyawake, and tapa) of grid size 9$\times$9 for sudoku, and 10$\times$10 for the rest. We use the validation set to track performance during training and select the final checkpoint. We report per-puzzle accuracy on five of these types on the golden set (TRM already reaches 100\% on shakashaka, so we omit it from the reported results), with aggregate scores sample-weighted across types.
We also report results on the \emph{Sudoku-Extreme}, \emph{Maze-Hard}, and \emph{ARC-AGI 2} datasets.

\textbf{Models and inference.} For each benchmark we use a standard TRM checkpoint. For
Sudoku-Extreme we use the TRM-MLP variant (which the TRM paper showed to be stronger on Sudoku), and for the other datasets, we use TRM-Att. PTRM inference uses $K$ parallel rollouts each running $D$ supervision steps with Gaussian noise of scale $\sigma$ added to the
latent state at each supervision step. The selected configuration $(K, D, \sigma)$ varies by benchmark and is given alongside each result. Metrics are averaged across three seeds.

\textbf{Baselines.} To isolate the contribution of PTRM's stochastic rollouts from the underlying backbone, we report standard TRM performance (the same checkpoint as PTRM ran deterministically). For each dataset, we report the performance of frontier LLMs. For Sudoku-Extreme, Maze-Hard, and ARC2 we additionally report the published direct prediction and TRM baselines from \citep{jolicoeur2025less}.

\textbf{Cost estimation.} PPBench provides the dollar cost per attempt for each LLM. We convert PTRM's wall-clock to a comparable dollar figure using a single H100 at \$2.50/hr (standard cloud pricing \citep{vastai_h100_pcie_pricing}) so that $\text{cost} = \$2.50 \cdot t_{\text{puzzle}} / 3600$, where $t_{\text{puzzle}}$ is the time (in seconds) to complete a puzzle.

\subsection{Pencil Puzzle Bench}
\label{sec:ppbench}

\subsubsection{Per-puzzle accuracy}
Table~\ref{tab:ppbench-main} reports per-puzzle accuracy on the PPBench golden set. PTRM at $K{=}100, D{=}48, \sigma{=}0.2$ raises aggregate best-Q@$K$ from $62.6\%$ to $91.2\%$. Increasing supervision depth alone ($K{=}1, D{=}48$) gives a small boost over the standard TRM baseline ($K{=}1, D{=}16$), and even $D{=}1600$, which matches the compute of $K{=}100$ rollouts at $D{=}16$, reaches only $69.4\%$. Most of the gain comes from scaling width (stochastic rollouts). The largest improvements are on puzzle types where the deterministic baseline performed the worst (most headroom): sudoku improves from $46.7\%$ to $97.8\%$ and tapa from $40.0\%$ to $80.0\%$.

\begin{table}[h]
\centering
\small
\setlength{\tabcolsep}{4pt}
\begin{tabular}{lcccccc|c}
\toprule
  \% accuracy & \# Params & sudoku & lightup & nurikabe & heyawake & tapa & \textbf{agg.} \\
\midrule
Direct prediction              & 27M & $0.0$  & $0.0$  & $0.0$  & $14.3$ & $0.0$ & $2.0$ \\
TRM ($K{=}1, D{=}16$)          & 7M  & $46.7$ & $87.5$ & $74.1$ & $\mathbf{85.7}$ & $40.0$ & $62.6$ \\
TRM ($K{=}1, D{=}48$)          & 7M  & $57.8$ & $87.5$ & $74.1$ & $\mathbf{85.7}$ & $40.0$ & $66.0$ \\
TRM ($K{=}1, D{=}1600$)          & 7M  & $60.0$ & $\mathbf{100}$ & $77.8$ & $\mathbf{85.7}$ & $40.0$ & $69.4$ \\
\rowcolor{rowhighlight}
PTRM, best-Q@$K$ ($K{=}100, D{=}16$) & 7M  & $93.3$ & $\mathbf{100}$ & $\mathbf{88.9}$ & $\mathbf{85.7}$ & $\mathbf{80.0}$ & $89.8$ \\
\rowcolor{rowhighlight}
PTRM, best-Q@$K$ ($K{=}100, D{=}48$) & 7M  & $\mathbf{97.8}$ & $\mathbf{100}$ & $\mathbf{88.9}$ & $\mathbf{85.7}$ & $\mathbf{80.0}$ & $\mathbf{91.2}$ \\
\bottomrule
\end{tabular}\vspace{2pt}
\caption{\textbf{PPBench per-puzzle accuracy on the golden set.} PTRM uses the same backbone as the deterministic TRM. Scaling depth alone to $D{=}1600$ (matches compute of $K{=}100$ rollouts at $D{=}16$) lifts aggregate accuracy by $6.8$ points over the standard $D{=}16$ baseline. Combining depth with $K{=}100$ stochastic ($\sigma{=}0.2$) rollouts raises accuracy by $28.6$ percentage points overall. The direct-prediction baseline is a larger transformer trained on the same data.}
\label{tab:ppbench-main}
\end{table}

\subsubsection{Comparison with frontier LLMs on golden set}
\label{subsec:ppbench_llms}
PPBench reported per-puzzle results for several frontier LLMs using two strategies: 1) direct response from a single prompt, and 2) multi-turn agentic strategy with verification. We report results for \emph{direct} and \emph{any} (best of any strategy attempted, including agentic). The agentic strategy gives the LLM substantially more
resources than PTRM has access to. It provides the LLM the ability to iteratively verify each move with a perfect verifier. The direct strategy is the fairer comparison since, while it may use the model provider's reasoning harness, it does not have direct access to a multi-turn verifier (the LLM could still self-verify by writing verification code within the same response). We additionally observe that the agentic strategy was applied selectively in the published PPBench data: across the LLMs we compare against, only $9.6\%$ of direct failures on the golden set were retried with agentic. We restrict the comparison to the 7 strongest LLMs that attempted every puzzle in our golden set: \texttt{claude-opus-4-6@thinking}, \texttt{gpt-5.2@xhigh}, \texttt{gemini-3.1-pro}, \texttt{gpt-5.2@high}, \texttt{claude-sonnet-4-6@thinking}, \texttt{gpt-5.2@medium}, and \texttt{kimi-k2.5}. Table~\ref{tab:ppbench-llms} lists the top 3 in each strategy block.

We additionally report an \emph{ensemble} score formed from these 7 LLMs where a puzzle counts as solved if at least one of them solved it via any strategy. This ensemble setup is deliberately stacked against PTRM. It assumes a perfect verifier since, if any of the 7 LLMs produced a correct answer under any strategy, the ensemble counts it as solved, even though in practice we would not have access to an oracle verifier. Although it is not deployable, we include the ensemble to demonstrate that even under these heavily favorable conditions, frontier LLMs fall well short of PTRM. Ensemble cost-per-attempt averages over the attempts of all 7 models on each puzzle, and cost-per-correct
divides total cost by the number of puzzles the ensemble solved.

Table~\ref{tab:ppbench-llms} reports the comparison. PTRM exceeds the strongest single LLM (direct strategy) by $56.5$ points aggregate ($91.2\%$ vs.\ $34.7\%$), and exceeds the LLM ensemble by $36$ points ($91.2\%$ vs.\ $55.1\%$) despite the ensemble's stacked advantages. PTRM's estimated cost per attempt is orders of magnitude lower, though the two are only roughly comparable, and PTRM is trained on these puzzle types, while the LLMs' training data is not public so we do not know whether they trained on similar puzzles.

\begin{table}[h]
\centering
\small
\setlength{\tabcolsep}{4pt}
\begin{tabular}{lccccc|c|cc}
\toprule
\% accuracy & sudoku & lightup & nurikabe & heyawake & tapa & \textbf{agg.} & \textbf{\$/att.} & \textbf{\$/corr.} \\
\midrule
\multicolumn{9}{c}{\emph{Direct}} \\
\midrule
gemini-3.1-pro                & $6.7$ & $75.0$ & $22.2$ & $0.0$ & $30.0$ & $24.5$ & $\$0.40$ & $\$1.62$ \\
gpt-5.2@xhigh                 & $20.0$ & $50.0$ & $0.0$ & $0.0$ & $50.0$ & $24.5$ & $\$1.79$ & $\$7.29$ \\
claude-opus-4-6@thinking      & $0.0$ & $87.5$ & $44.4$ & $0.0$ & $60.0$ & $34.7$ & $\$2.91$ & $\$8.40$ \\
\midrule
\multicolumn{9}{c}{\emph{Any strategy (direct or agentic)\textsuperscript{$\dagger$}}} \\
\midrule
gemini-3.1-pro                & $6.7$ & $87.5$ & $33.3$ & $0.0$ & $40.0$ & $30.6$ & $\$10.38$ & $\$33.91$ \\
gpt-5.2@xhigh                 & $33.3$ & $75.0$ & $0.0$ & $0.0$ & $60.0$ & $34.7$ & $\$3.09$ & $\$8.90$ \\
claude-opus-4-6@thinking      & $0.0$ & $87.5$ & $44.4$ & $0.0$ & $70.0$ & $36.7$ & $\$4.38$ & $\$11.92$ \\
\midrule
\multicolumn{9}{c}{\emph{LLM ensemble\textsuperscript{$\dagger$}}} \\
\midrule
Any strategy (direct or agentic)                 & $46.7$ & $\mathbf{100}$ & $44.4$ & $0.0$ & $\mathbf{80.0}$ & $55.1$ & $\$2.66$ & $\$38.51$ \\
\midrule
\multicolumn{9}{c}{\emph{Ours, trained from scratch, 7M parameters}} \\
\midrule
\rowcolor{rowhighlight}
PTRM, best-Q@K & $\mathbf{97.8}$ & $\mathbf{100}$ & $\mathbf{88.9}$ & $\mathbf{85.7}$ & $\mathbf{80.0}$ & $\mathbf{91.2}$ & $\mathbf{\$0.001}$ & $\mathbf{\$0.001}$ \\
\bottomrule
\end{tabular}
\caption{\textbf{PTRM vs. frontier LLMs on PPBench golden.}
Per-puzzle accuracy and per-attempt / per-correct cost on the golden set. LLM costs are from PPBench. PTRM cost is estimated from H100 wall-clock (Sec. \ref{sec:setup}). The direct and agentic blocks list the 3 highest scoring LLMs on aggregate, and the ensemble row uses all 7 listed in Sec. \ref{subsec:ppbench_llms}. \textsuperscript{$\dagger$}Assumes access to a perfect verifier.}
\label{tab:ppbench-llms}
\end{table}

\subsection{Sudoku-Extreme, Maze-Hard, and ARC-AGI-2}
\label{sec:other-benchmarks}

For each benchmark we use the standard TRM checkpoint trained as described in \citep{jolicoeur2025less} without modification (TRM-MLP for Sudoku-Extreme and TRM-Att for Maze-Hard and ARC-AGI-2). Table~\ref{tab:other-benchmarks} summarizes results on all three.

On Sudoku-Extreme, PTRM at $K{=}100, D{=}64, \sigma{=}0.3$ raises the deterministic baseline of $87.3\%$ to $99.06\%$ pass@$K$ and $98.75\%$ best-Q@$K$, achieving state of the art.

On Maze-Hard, PTRM at $K{=}100$, $D{=}16$, $\sigma{=}1.0$ reaches $95.73\%$ pass@$K$, an $11.93$ point gain over the $83.8\%$ deterministic baseline. mode@$K$ gives the best PTRM accuracy here at $87\%$ ($+3.2$ points), with best-Q@$K$ behind at $84.5\%$ ($+0.7$ points). While pass@$K$ shows that PTRM is able to unlock several correct answers, the Q head identifies them less reliably than on the previous benchmarks. Following the standard for Maze-Hard, these numbers are when comparing the final answers with labels from the dataset. Mazes can have many possible shortest paths, but those labels represent only one. When accepting any shortest path, TRM reaches $90.8\%$, mode@$K$ $93.2\%$, best-Q@$K$ $91.3\%$ and pass@$K$ $98.0\%$ (Appendix~\ref{app:maze-valid}).

On ARC-AGI-2, the standard inference pipeline applies data augmentations and votes across them. PTRM adds $K$ stochastic rollouts per augmentation. For selection, we pick the rollout with the highest Q value within each augmentation, then vote across augmentations as in the standard pipeline. With $K{=}25$ and $\sigma{=}0.2$, PTRM lifts pass@1 from $7.36\%$ to $8.47\%$ and pass@100 from $14.31\%$ to $15.97\%$ over our deterministic TRM baseline, while matching it at pass@2.

\begin{table}[h]
\centering
\small
\setlength{\tabcolsep}{4pt}
\begin{tabular}{lccccccc}
\toprule
 &           & Sudoku-Extreme & Maze-Hard & \multicolumn{3}{c}{ARC-AGI-2} \\
Method & \# Params & Acc.\ (\%) & Acc.\ (\%) & pass@1 & pass@2 & pass@100 \\
\midrule
HRM                                    & 27M  & $55.0$           & $74.5$           & --   & $\phantom{0}5.0$ & --   \\
TRM                                & 5M / 7M\textsuperscript{$\dagger$}   & $87.4$           & $85.3$           & --   & $\phantom{0}7.8$ & --   \\
\midrule
\multicolumn{7}{l}{\emph{Ours}} \\
Standard TRM, our reproduction         & 5M / 7M\textsuperscript{$\dagger$} & $87.28$           & $83.80$           & $\phantom{0}7.36$           & $\phantom{0}\mathbf{9.72}$           & $14.31$           \\
\rowcolor{rowhighlight}
PTRM                       & 5M / 7M\textsuperscript{$\dagger$} & $\mathbf{98.75}$           & $\mathbf{87.00}$           & $\phantom{0}\mathbf{8.47}$  & $\phantom{0}\mathbf{9.72}$  & $\mathbf{15.97}$  \\
\bottomrule
\end{tabular}

\caption{\textbf{Sudoku-Extreme, Maze-Hard, and ARC-AGI-2 results.}
For Sudoku-Extreme, $K{=}100$, $D{=}64$, $\sigma{=}0.3$. For Maze-Hard, $K{=}100$, $D{=}16$, $\sigma{=}1.0$. For ARC-AGI-2, $K{=}25$, $D{=}16$, $\sigma{=}0.2$. pass@$k$ for ARC-AGI-2 reports the top-$k$ predictions from the augmentation-voting pipeline. PTRM shows an accuracy improvement over standard TRM across all 3 benchmarks. \textsuperscript{$\dagger$}Following~\citep{jolicoeur2025less}, 5M for Sudoku-Extreme (TRM-MLP), 7M for Maze-Hard and ARC-AGI-2 (TRM-Att).} 
\label{tab:other-benchmarks}
\vspace{-6ex}
\end{table}

\subsection{Multi-answer puzzles: N-Queens}
\label{sec:nqueens}
Most of the previous benchmarks have only one correct answer per puzzle. To test PTRM on puzzles with several possible answers, we use N-Queens where, given an $N{\times}N$ board with some queens that are not attacking each other, the task is to place the rest of the queens such that no two queens attack each other. We train TRM-Att from scratch on $10{\times}10$ and $16{\times}16$ boards only using puzzles with exactly one answer and evaluate on held-out puzzles which may have many. The direct-prediction baseline is trained on the same data (Appendix~\ref{app:nqueens}).

\begin{table}[h]
\centering
\small
\setlength{\tabcolsep}{6pt}
\begin{tabular}{lccc}
\toprule
\% accuracy & \# Params & $10{\times}10$ & $16{\times}16$ \\
\midrule
Direct prediction & 27M & $30.2$ & $14.4$ \\
TRM ($K{=}1, D{=}16$) & 7M & $82.4$ & $48.1$ \\
TRM ($K{=}1, D{=}320$) & 7M & $89.0$ & $68.8$ \\
PTRM, mode@$K$ ($K{=}20, D{=}16$) & 7M & $80.9$ & $43.1$ \\
\rowcolor{rowhighlight}
PTRM, best-Q@$K$ ($K{=}20, D{=}16$) & 7M & $\mathbf{94.9}$ & $\mathbf{75.5}$ \\
\bottomrule
\end{tabular}\vspace{2pt}
\caption{\textbf{N-Queens results.} Valid completions on held-out inputs (most of which have several correct answers). PTRM uses $\sigma{=}1.0$. TRM at $D{=}320$ uses the same compute as PTRM's 20 rollouts.}
\label{tab:nqueens}
\end{table}

As shown in Table~\ref{tab:nqueens}, best-Q@$K$ raises accuracy from $82.4\%$ to $94.9\%$ on $10{\times}10$ and from $48.1\%$ to $75.5\%$ on $16{\times}16$, beating depth scaling at matched-compute and significantly outperforming mode@$K$. PTRM is also able to find different solutions to the same puzzle. Its rollouts cover on average $77\%$ ($10{\times}10$) and $53\%$ ($16{\times}16$) of each input's valid completions, against $50\%$ and $40\%$ for standard TRM at $D{=}320$ (Appendix~\ref{app:nqueens}).

\subsection{Exploration and selection}
\label{sec:experiments-discussion}

With more noise (higher $\sigma$), PTRM finds many correct solutions that the
deterministic inference misses. For instance, on Maze-Hard, the deterministic model solves 83.8\% of puzzles, but PTRM raises pass@$K$ to nearly 96\%. The extent to which PTRM helps depends on the task, but on every dataset we tested, it unlocks correct solutions well beyond the deterministic model's reach.

We find that early noise (particularly at the initial step) drives most of the gains and simply adding noise at every step works best compared to other noise schedules. Sampling the output instead of the latent state finds nearly no new correct answers. Appendix~\ref{app:noise-ablation} reports a full noise $\sigma$ sweep, as well as noise schedule and sampling method ablations.

TRM's jointly trained Q head serves as a strong verifier on most tasks. On PPBench and Sudoku-Extreme, best-Q@$K$ reaches values within about a point of the pass@$K$, so PTRM's exploration translates directly into accuracy gains. On N-Queens, best-Q@$K$ significantly outperforms majority voting (mode@$K$). In Maze-Hard, more exploration (higher $\sigma$) produces significantly more correct rollouts, but even taking into account the fact that a maze can have multiple shortest paths, the existing Q head is not able to identify them, leaving performance on the table.

\noindent
\begin{minipage}[c]{0.47\linewidth}
\textbf{Early stopping.} PTRM does not always need all full $K$ rollouts. Since the Q head estimates the correctness of an answer, we can stop sampling as soon as it is confident in one ($\mathrm{sigmoid}(\hat{q}) > \tau$). On the PPBench validation set, this mechanism matches the accuracy of standard PTRM ($K{=}100$ rollouts) with about $20\times$ less compute (Figure~\ref{fig:halting}, Appendix~\ref{app:inference-cost}). We leave better selection criteria and halting mechanisms to future work.
\end{minipage}\hfill
\begin{minipage}[c]{0.50\linewidth}
\centering
\includegraphics[width=\linewidth]{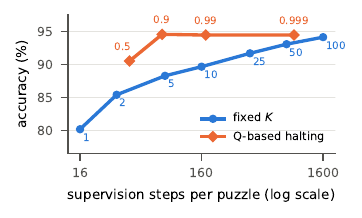}
\captionof{figure}{\textbf{Q-based halting.} On PPBench validation. Labels: number of rollouts $K$ (blue), halting confidence threshold $\tau$ (orange).}
\label{fig:halting}
\end{minipage}

\section{Related Work}\label{sec:rw}
A long line of work explores recursive computation for iterative reasoning and representation refinement. Early examples include Universal Transformers \citep{dehghani2018universal}, Mixture-of-Recursions \citep{bae2025mixture}, Deep Thinking models \citep{schwarzschild2021can, bansal2022end, bear2024rethinking}, and HRM \citep{wang2025hierarchical}, all of which investigate the use of repeated computation steps to improve reasoning performance. More recent work has introduced methods to substantially accelerate TRM training \citep{hakimi2026form}, while TRM-style recursive architectures have also been extended to language modeling tasks \citep{li2026learning}.

Building on this broader perspective of recursive computation, a growing body of work studies latent-space reasoning through the reuse of hidden states. \citet{hao2024training} propose continuous “thinking tokens” derived from Chain-of-Thought (CoT) traces \citep{wei2022chain}, which are autoregressively generated and appended to the model context, enabling reasoning directly in latent space without producing intermediate textual outputs. Similarly, \citet{zhu2025reasoning} formalize learning by superposition and demonstrate improvements on tasks such as graph reachability. By avoiding explicit token sampling and implicitly representing multiple reasoning trajectories, these approaches may mitigate the unfaithfulness and backtracking often observed in standard autoregressive reasoning \citep{lanham2023measuring, chen2025reasoning}.



Related to our work, \citet{baekgenerative} propose a generative version of TRM where the hidden state $z$ is sampled instead of deterministic. This improves performance on multiple tasks, but requires retraining. In a concurrent proof-of-concept study on a subset of Sudoku-Extreme, \citet{efstathiou2026recursive} also add noise at test time to the initial hidden state only.

Our observations in Sec.~\ref{sec:basins} are consistent with the mechanistic analysis of \citet{ren2026your}, who identify spurious fixed points in HRM’s latent dynamics on Sudoku-Extreme. Their method mitigates these attractors through a combination of task-specific training data augmentation, inference-time input perturbations, and model bootstrapping across training checkpoints, thereby effectively increasing test-time compute. However, these interventions are comparatively less general and less computationally efficient. In contrast, we observe analogous basin structure in TRM across multiple puzzle types and achieve attractor escape using a substantially simpler, task-agnostic mechanism: injecting Gaussian noise into the latent state at each supervision step while using a single deterministic checkpoint.

\section{Conclusion}\label{sec:conclusion}
In this work, we introduced Probabilistic TRM (PTRM), a novel test-time scaling paradigm for Tiny Recursive Models (TRM) through parallel exploration and selection. This approach scales test-time compute using \emph{width} ($K$ parallel rollouts), yielding larger gains than \emph{depth} scaling (increasing deep recursion steps) alone. PTRM requires no retraining and does not rely on task-specific data augmentations making it extremely easy to use and versatile.

By scaling both width and depth, PTRM obtains significant gains in accuracy when tested on a wide selection of puzzles. On PPBench (Sudoku, Lightup, Nurikabe, Heyawake, Tapa puzzles), PTRM nearly obtains twice the accuracy ($91.2\%$) of an ensemble of SOTA LLMs ($55.1\%$) at a fraction of the cost. Furthermore, PTRM improves accuracy on Sudoku (from $87.28\%$ to $98.75\%$), Maze-Hard (from $83.80\%$ to $87.00\%$), and ARC-AGI (from $7.36\%$ to $8.47\%$ pass@1). On N-Queens, PTRM is able to find multiple correct solutions for the same puzzle.

\textbf{Limitations.} Our experiments focus on reasoning puzzles rather than general tasks. We only test on a subset of PPBench puzzles. We are limited to puzzles with a small grid-size due to limited computational resources. It is not guaranteed that the method works as well for all types of problems (e.g., accuracy gains on ARC-AGI-2 and Heyawake are smaller).

\textbf{Future work.} It would be interesting to understand why some puzzles benefit from test-time scaling more than others. We suspect that problems that are harder to verify (e.g., ARC-AGI-2) benefit less from PTRM because the Q head may struggle to distinguish correct solutions from incorrect ones. Developing stronger verifiers than the existing Q head, as well as improving the selection and halting mechanisms are interesting directions for future work.

\begin{ack}
This research was enabled in part by compute resources provided by Mila and the Digital Research Alliance of Canada. We thank Marco Pedersoli and Pablo Piantanida for their financial support and for providing access to computational resources. Ali Parviz was funded by NSF CIF-2403452. We declare no competing interests.
\end{ack}

\bibliographystyle{unsrtnat}
\bibliography{ref}

\newpage
\appendix

\section{Implementation Details}
\label{app:implementation-details}

\subsection{Compute}
\label{app:compute}

We train and evaluate all models on a single NVIDIA H100 80GB GPU. PTRM introduces no additional training cost over standard TRM since it operates entirely at inference time.

\subsection{Models}
\label{app:models}

All experiments use the standard TRM backbone~\citep{jolicoeur2025less} with the released architecture and training recipes. Following the TRM paper, we use the MLP variant (\emph{TRM-MLP}, 5M parameters) for Sudoku-Extreme and the attention variant (\emph{TRM-Att}, 7M parameters) for Maze-Hard, ARC-AGI-2, and PPBench. Layout and hyperparameters are unchanged from TRM.

\subsection{PPBench dataset construction}
\label{app:ppbench-data}

Sudoku-Extreme, Maze-Hard, and ARC-AGI-2 use the same checkpoints and data
splits as TRM. The PPBench dataset is more recent and has previously been used only with frontier LLMs, so we detail how we built our training, validation, and golden splits.

\paragraph{Source.}
PPBench contains $62{,}231$ constraint-satisfaction pencil puzzles spanning 94 puzzle types. Of these, 300 puzzles ($15$ puzzles $\times$ $20$ types) are held out as the \emph{golden} benchmark set by Waugh~\citep{waugh2026pencil}.

\paragraph{Filtering.}
From the remaining $61{,}931$ puzzles we hold out a validation set by sampling 100 puzzles from each puzzle type (50 for \texttt{tapa}, due to its smaller base size), and the rest forms the training set. We then filter all three sets (training, validation, golden) to retain only puzzles of six types (\texttt{sudoku}, \texttt{lightup}, \texttt{nurikabe}, \texttt{shakashaka},
\texttt{heyawake}, \texttt{tapa}) at fixed grid sizes: $9{\times}9$ for
sudoku and $10{\times}10$ for the others. Sudoku grids are padded with a
pad token to $10{\times}10$, giving a uniform sequence length of $\text{seq\_len}=100$ across all six puzzle types. The deterministic TRM baseline reaches 100\% accuracy on shakashaka, so we exclude it from per-puzzle accuracy reporting (no headroom to compare against PTRM).

\paragraph{Augmentation.}
Each training puzzle is expanded into 10 examples using two augmentations:
1) \emph{trajectory sampling}, where the input is set to a random intermediate
solve state along the puzzle's solution trajectory rather than always the empty
initial grid, while the label is always the fully solved grid; and
2) \emph{dihedral transformation}, where a random dihedral transformation of a square grid, among the 8 possibilities given by 4 rotations $\times$ 2 \{identity, reflection\}, is applied to both the input and the label.
For each puzzle, the first example is the unaugmented (initial state,
solved) pair. The remaining 9 are randomly sampled (trajectory and dihedral transform). Validation and golden splits are not augmented.

\paragraph{Resulting splits.}
The merged multi-type splits use a unified vocabulary of 294 tokens and
$\text{seq\_len}=100$. Per-type sample counts are reported in
Table~\ref{tab:ppbench-splits}.

\begin{table}[h]
\centering
\small
\setlength{\tabcolsep}{8pt}
\begin{tabular}{lrrr}
\toprule
puzzle type & train & val & golden \\
\midrule
sudoku                              & $\phantom{0}7{,}810$  & $\phantom{0}97$ & $15$ \\
lightup                             & $\phantom{0}9{,}504$  & $\phantom{0}65$ & $\phantom{0}8$ \\
nurikabe                            & $15{,}180$            & $\phantom{0}55$ & $\phantom{0}9$ \\
heyawake                            & $42{,}108$            & $\phantom{0}70$ & $\phantom{0}7$ \\
tapa                                & $\phantom{0}3{,}663$  & $\phantom{0}26$ & $10$ \\
shakashaka\textsuperscript{$\ast$}  & $20{,}702$            & $\phantom{0}62$ & $12$ \\
\midrule
total                               & $98{,}967$            & $375$           & $61$ \\
\bottomrule
\end{tabular}
\caption{Per-puzzle-type sample counts in the PPBench splits used in
training and evaluation. \textsuperscript{$\ast$}Shakashaka is included in training but excluded from per-puzzle accuracy reporting because deterministic TRM already solves all evaluated shakashaka puzzles.}
\label{tab:ppbench-splits}
\end{table}

\section{Noise Ablation}
\label{app:noise-ablation}

\begin{figure}[h]
\centering
\includegraphics[width=\linewidth]{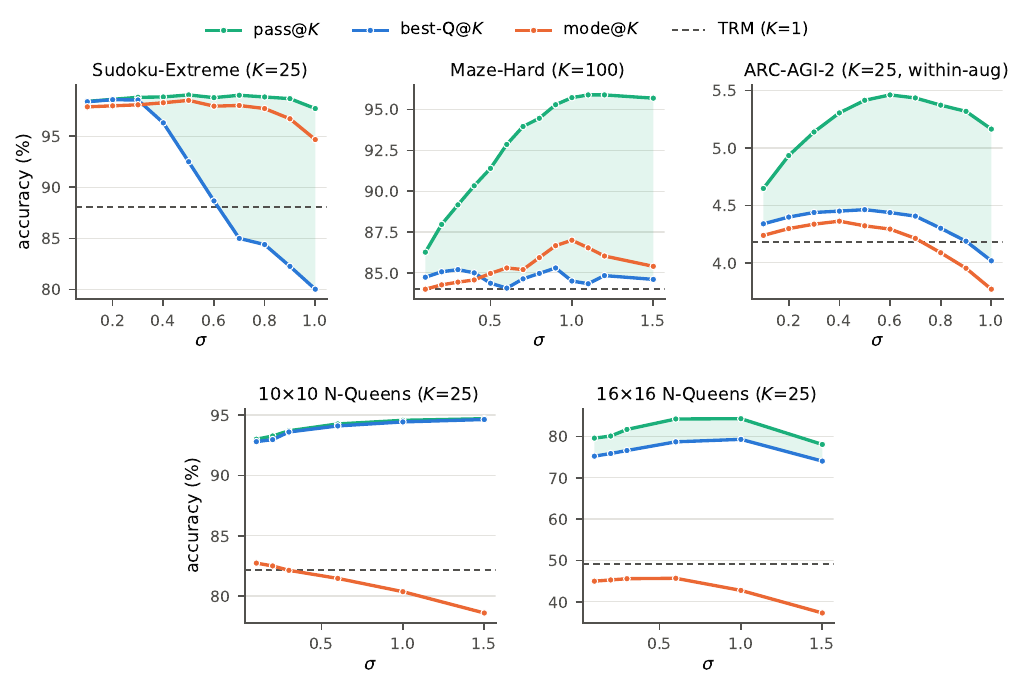}
\caption{\textbf{pass@$K$, best-Q@$K$, and mode@$K$ across $\sigma$ per rollout batch.}
On every task, increasing the inference noise consistently produces more correct rollouts (pass@K, green) up to a task-dependent $\sigma$ value. The Q head (best-Q@$K$, blue) tracks the pass@$K$ ceiling closely on Sudoku-Extreme (at low $\sigma$) and $10{\times}10$ N-Queens, and leaves a larger gap on Maze-Hard, ARC-AGI-2, and $16{\times}16$ N-Queens. The shaded region represents the verifier headroom (accuracy that a better verifier could extract). mode@$K$ (orange) has the edge over the Q head only on Maze-Hard. For ARC-AGI-2, metrics are per puzzle/augmentation to isolate the Q head's verification abilities from the augmentation pipeline.}
\label{fig:noise-ablation}
\end{figure}

We ablate the inference noise level $\sigma$ on five benchmarks at $K{=}25$ ($K{=}100$ for Maze-Hard) and $D{=}16$ to keep the sweep tractable: Sudoku-Extreme, Maze-Hard, ARC-AGI-2, and $10{\times}10$ and $16{\times}16$ N-Queens. For Sudoku-Extreme and N-Queens (per board size), to further limit computational cost, we randomly sample 1000 puzzles from the test set. Figure~\ref{fig:noise-ablation} shows pass@$K$, best-Q@$K$, and mode@$K$ as a function of $\sigma$, averaged over three random seeds. mode@$K$ selects the most frequent answer among the $K$ rollouts (ties go to the earliest rollout). On N-Queens, any valid completion of the input counts as a correct answer.

On Maze-Hard pass@$K$ climbs from 83.8\% (deterministic) to nearly 96\% by $\sigma{\approx}1.0$ and then plateaus. On Sudoku-Extreme it is already near its ceiling at $\sigma{=}0.1$ and stays roughly flat across the sweep. On ARC-AGI-2 it peaks near $\sigma{=}0.6$ before declining. Q head selection nearly matches the ceiling (maximum pass@$K$) on Sudoku-Extreme while best-Q@$K$ peaks at $98.5\%$ (within a point of pass@$K$'s peak of $99\%$). On the other hand, the gap between best-Q@$K$ and maximum pass@$K$ is more pronounced on Maze-Hard and ARC-AGI-2 (headroom a stronger verifier could close).

Noise affects best-Q@$K$ and mode@$K$ differently. On Sudoku-Extreme, best-Q@$K$ falls from $98.5\%$ at $\sigma{\le}0.3$ to $80.0\%$ at $\sigma{=}1.0$, while mode@$K$ stays within 4 points of its $98.5\%$ peak. On N-Queens, mode@$K$ is at best half a point above the deterministic baseline and falls below it as $\sigma$ grows. Valid rollouts spread across different solutions while incorrect ones are often identical, so the single most common output is often an invalid one (Appendix~\ref{app:nqueens}). best-Q@$K$ does not suffer from the same failure mode since the Q head scores each rollout independently, gaining $12.4$ points on $10{\times}10$ and $30.2$ points on $16{\times}16$.

\paragraph{Noise schedule.} There are many ways to inject noise at inference. We ablate different noise schedules in Table~\ref{tab:noise-schedule} on the PPBench validation set ($K{=}25$, $D{=}16$, $\sigma{=}0.3$, 3 seeds). Noise in the early steps, especially the initial one, is what matters most. Schedules that inject noise early are within about 3 points of each other, while schedules that do not initially add noise (linear growth and second half only) lose about 6 points. We use the every-step noise schedule throughout the paper since it performs best and produces the most distinct answers.

\begin{table}[h]
\centering
\small
\setlength{\tabcolsep}{5pt}
\begin{tabular}{lccc}
\toprule
Noise schedule & best-Q@25 & pass@25 & distinct answers \\
\midrule
every step & $\mathbf{91.8}\pm0.8$ & $\mathbf{94.4}\pm0.2$ & $9.3$ \\
linear decay & $90.2\pm0.9$ & $93.7\pm0.7$ & $8.4$ \\
cosine decay & $90.0\pm1.7$ & $93.4\pm1.1$ & $8.4$ \\
first half only & $89.9\pm1.1$ & $93.7\pm0.7$ & $8.5$ \\
initialization only & $89.6\pm0.6$ & $93.6\pm0.7$ & $8.0$ \\
every 4th step & $89.1\pm0.9$ & $93.3\pm0.0$ & $8.2$ \\
linear growth & $86.0\pm0.5$ & $86.3\pm0.5$ & $7.5$ \\
second half only & $85.6\pm0.5$ & $85.9\pm0.3$ & $7.3$ \\
\bottomrule
\end{tabular}\vspace{2pt}
\caption{\textbf{Noise schedules.} On PPBench validation ($K{=}25$, $D{=}16$, $\sigma{=}0.3$, mean $\pm$ std over 3 seeds). Decay and growth schedules scale $\sigma$ over the 16 steps. \textit{Distinct answers} represents the mean number of different outputs among the 25 rollouts.}
\label{tab:noise-schedule}
\end{table}

\paragraph{Output sampling.} For LLMs, a standard way of scaling test-time compute is to sample at the output level \citep{snell2024scaling}. The analogue for TRM would be to sample each token of the final answer from its output distribution instead of taking the argmax. Table~\ref{tab:output-sampling} reports the best results of a temperature sweep, showing that output sampling adds almost nothing (pass@$K$ barely changes). Since the final answer tokens are sampled independently, samples predominantly either reproduce the argmax answer or turn into invalid solutions. Moreover, the Q head output is computed from the latent state; since it cannot see the sampled outputs, it gives every sample the same score. Latent noise instead allows the model to explore different trajectories that each end in a coherent answer with its own Q score.

\begin{table}[h]
\centering
\small
\begin{tabular}{lccc}
\toprule
 & TRM & Output sampling, pass@$K$ & PTRM, pass@$K$ \\
\midrule
PPBench val (313), $K{=}100$ & $81.2$ & $81.5$ & $\mathbf{96.0}$ \\
Sudoku-Extreme (1000), $K{=}25$ & $88.9$ & $89.0$ & $\mathbf{98.4}$ \\
\bottomrule
\end{tabular}\vspace{2pt}
\caption{\textbf{Output sampling vs. latent noise.} Output sampling uses the best temperature per benchmark.}
\label{tab:output-sampling}
\end{table}

\section{Inference cost}
\label{app:inference-cost}
\begin{figure}[h]
\centering
\includegraphics[width=\linewidth]{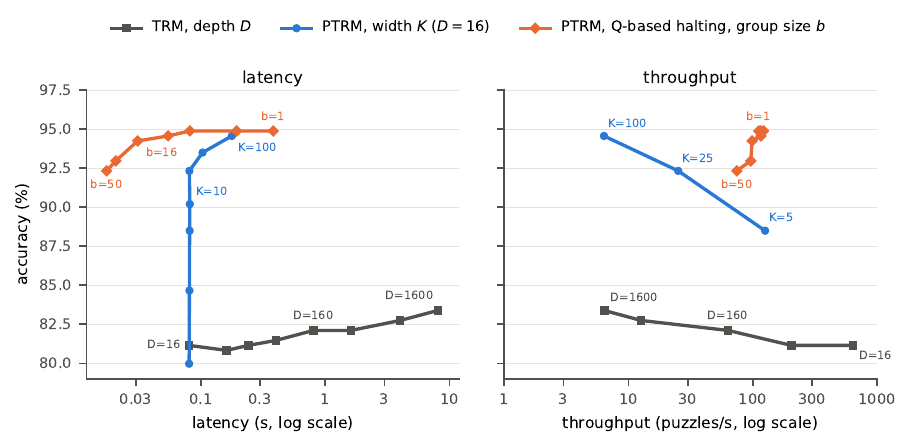}
\caption{\textbf{Accuracy against latency and throughput.} On PPBench validation (using a single H100 GPU). Throughput uses a batch size of $1024$. Q-based halting uses $\tau{=}0.9$.}
\label{fig:cost}
\end{figure}

Unlike depth scaling, width scaling can perform rollouts independently and in parallel, so, even at matched compute, width can be faster than depth. Moreover, PTRM can stop early with Q-based halting (Sec.~\ref{sec:experiments-discussion}), where a group of $b$ rollouts are run in parallel for each puzzle, and the final answer is returned when any of the rollouts has $\mathrm{sigmoid}(\hat{q}) > \tau$, otherwise the next group of $b$ rollouts starts. This keeps the benefits from stochastic rollouts for a fraction of the cost. It works because, as shown in Figure~\ref{fig:nqueens-rollouts}, many rollouts converge to a valid solution, which the Q head can detect, within a few steps, so running every rollout for all D steps mostly wastes compute. Figure~\ref{fig:cost} and Table~\ref{tab:cost} measure depth, width, and Q-based halting on PPBench validation in terms of latency (the time to answer one puzzle at batch size $1$) and throughput (answered puzzles per second with a batch size of $1024$). For throughput, we continuously replace a finished rollout with the next one in the queue (continuous batching).

For latency, since we use a batch size of $1$, the setup highly favors methods that leverage parallel computation, such as PTRM. On an H100 GPU, up to $25$ rollouts can be performed in the same amount of time as one. $K{=}100$ rollouts answer in $0.18$\,s, against $8.0$\,s for depth scaling at matched compute ($D{=}1600$), while being around $11$ points more accurate. For throughput, we use a batch size of $1024$, which saturates the GPU (compute-bound regime). Depth then also leverages parallel compute by filling the batch with different puzzles, so parallelism no longer favors width. Throughput depends only on the steps spent per puzzle, so $K{=}100$ and $D{=}1600$ serve puzzles at the same rate. At equal throughput, width is still more accurate than depth.

\begin{table}[h]
\centering
\small
\begin{tabular}{lcccc}
\toprule
 & accuracy & steps & latency (s) & throughput (puzzles/s) \\
\midrule
TRM, $D{=}16$ & $81.2$ & $16$ & $0.08$ & $640$ \\
TRM, $D{=}1600$ & $83.4$ & $1600$ & $8.0$ & $6.4$ \\
PTRM, $K{=}100$ & $94.6$ & $1600$ & $0.18$ & $6.4$ \\
PTRM, Q-based halting, sequential ($b{=}1$) & $94.9$ & $76$ & $0.38$ & $112$ \\
PTRM, Q-based halting, $b{=}16$ & $94.2$ & $92$ & $0.03$ & $99$ \\
\bottomrule
\end{tabular}\vspace{2pt}
\caption{\textbf{Inference cost on PPBench validation.} On a single H100 GPU. \textit{Steps} represents supervision steps per puzzle. \textit{Latency} represents the time to answer one puzzle at batch size 1. \textit{Throughput} represents the number of puzzles answered per second (with batch size of $1024$). \textit{Halting} represents Q-based halting with $\tau{=}0.9$.}
\label{tab:cost}
\end{table}

With Q-based halting ($\tau{=}0.9$), most puzzles are answered long before all $100$ rollouts finish. Running rollouts sequentially ($b{=}1$) matches the accuracy of $K{=}100$ while only spending $76$ supervision steps per puzzle on average instead of $1600$. For throughput, this translates into about $18\times$ more puzzles per second, and a larger group size ($b$) does not help. For latency, larger groups help since, with a batch size of $1$, running more rollouts in parallel adds almost no latency while making it more likely that one of them becomes confident within the first few steps. With $b{=}16$, halting answers in $0.03$\,s at $94.2\%$, about $6\times$ faster than standard PTRM with all $K{=}100$ rollouts ($0.18$\,s, $94.6\%$). However, larger groups lose some accuracy, as they are more likely to stop on a rollout that is confident early but wrong ($92.3\%$ at $b{=}50$). Halting behaves similarly on N-Queens where, at $\tau{=}0.9$, it matches standard PTRM ($K{=}20$) at $\sim95\%$ on $10{\times}10$ and $\sim75\%$ on $16{\times}16$, while using far fewer steps ($28$ steps on $10{\times}10$ and $79$ on $16{\times}16$, instead of $320$).

\section{Per-puzzle accuracy on the PPBench validation set}
\label{app:ppbench-val}

The main paper reports per-puzzle accuracy on the PPBench golden set
(Table~\ref{tab:ppbench-main}) for direct comparability with the
LLM evaluations from Waugh~\citep{waugh2026pencil} who used that set. For a lower-variance complement, Table~\ref{tab:ppbench-val} reports results on our validation set ($313$ puzzles across the five reported types). Depth scaling alone gives at most a small lift ($81.2\%$ at $D{=}16$ and $83.4\%$ at $D{=}1600$), while stochastic rollouts ($K{=}100, D{=}16, \sigma{=}0.2$) raise aggregate best-Q@$K$ from $81.2\%$ to $94.6\%$, a $13.4$ percentage-point improvement. Adding depth on top ($D{=}48$) only helps sudoku and slightly lowers the aggregate score. The biggest gains are again on puzzles where the deterministic baseline has the most headroom (sudoku $62.9\%$ to $96.9\%$, tapa $57.7\%$ to $74.4\%$). Types where the baseline is already near ceiling (heyawake at $95.7\%$) increase only marginally.

\begin{table}[h]
\centering
\small
\setlength{\tabcolsep}{4pt}
\begin{tabular}{lcccccc|c}
\toprule
\% accuracy & \# Params & sudoku & lightup & nurikabe & heyawake & tapa & \textbf{agg.} \\
\midrule
TRM ($K{=}1, D{=}16$) & 7M & $62.9$ & $93.8$ & $90.9$ & $95.7$ & $57.7$ & $81.2$ \\
TRM ($K{=}1, D{=}1600$) & 7M & $74.2$ & $95.4$ & $89.1$ & $95.7$ & $42.3$ & $83.4$ \\
\rowcolor{rowhighlight}
PTRM, best-Q@$K$ ($K{=}100, D{=}16$) & 7M & $96.9$ & $\mathbf{96.9}$ & $\mathbf{93.9}$ & $\mathbf{97.1}$ & $\mathbf{74.4}$ & $\mathbf{94.6}$ \\
\rowcolor{rowhighlight}
PTRM, best-Q@$K$ ($K{=}100, D{=}48$) & 7M & $\mathbf{97.6}$ & $\mathbf{96.9}$ & $91.5$ & $\mathbf{97.1}$ & $66.7$ & $93.7$ \\
\bottomrule
\end{tabular}
\caption{\textbf{PPBench per-puzzle accuracy on the validation set.}
PTRM uses the same backbone as the deterministic TRM.}
\label{tab:ppbench-val}
\end{table}

\section{Puzzles with several valid answers}
\label{app:multi}

\subsection{N-Queens}
\label{app:nqueens}
\paragraph{Data.} Every training input has exactly one valid solution (checked with a solver), so the model never sees an input with several answers during training. $10{\times}10$ N-Queens has only 724 solutions. We form inputs by keeping 1--3 queens of a solution, which gives a set of distinct partial boards (where different boards can share a solution), and split that set 85:15 into training and test sets. We train on the 13,587 single-solution training inputs. The 6,513 test inputs can have many solutions. Test inputs are never seen in training, but since there are so few solutions, $99.5\%$ of their completions also appear as training targets. $16{\times}16$ has over 14 million solutions, which allows us to split by solution instead and provides a stronger test where no test solution is seen during training. Each input is built from a sampled solution, which serves as its target. We place the solution's queens on an empty board one at a time, in random order, until the partial board has only one valid completion. Some training inputs get 0--4 extra queens to make them easier. Training inputs come from 100,000 sampled solutions. Test inputs are built the same way from held-out solutions, after which we remove up to 4 of their queens at random so that they can have several completions. A validation set built the same way from a separate set of solutions is used to select the $16{\times}16$ checkpoint. We train with TRM-Att and the standard recipe. Table~\ref{tab:nqueens-full} breaks Table~\ref{tab:nqueens} down by the number of queens given (i.e., the number of pre-placed queens).

\begin{table}[h]
\centering
\small
\setlength{\tabcolsep}{2.5pt}
\begin{tabular}{llrrcc|ccc|c}
\toprule
Board & Queens & Inputs & Solutions & TRM $D{=}16$ & TRM $D{=}320$ & mode@20 & best-Q@20 & pass@20 & Distinct \\
\midrule
$10{\times}10$ & 1 & 15 & 76 & $93.3$ & $\mathbf{100.0}$ & $66.7$ & $\mathbf{100.0}$ & $100.0$ & $10.20$ \\
 & 2 & 534 & 9 & $87.3$ & $95.1$ & $77.3$ & $\mathbf{99.1}$ & $99.1$ & $3.97$ \\
 & 3 & 5,964 & 2 & $82.0$ & $88.4$ & $81.2$ & $\mathbf{94.6}$ & $94.8$ & $1.60$ \\
 & all & 6,513 & 2 & $82.4$ & $89.0$ & $80.9$ & $\mathbf{94.9}$ & $95.2$ & $1.81$ \\
\midrule
$16{\times}16$ & 1--4 & 266 & 718 & $0.4$ & $6.4$ & $0.0$ & $\mathbf{13.5}$ & $16.5$ & $0.22$ \\
 & 5 & 322 & 81 & $15.5$ & $50.0$ & $3.7$ & $\mathbf{66.8}$ & $79.2$ & $2.05$ \\
 & 6 & 396 & 12 & $30.3$ & $64.9$ & $11.6$ & $\mathbf{73.2}$ & $85.1$ & $2.34$ \\
 & 7 & 436 & 3 & $42.9$ & $72.0$ & $33.7$ & $\mathbf{77.5}$ & $85.8$ & $1.83$ \\
 & 8 & 406 & 1 & $61.1$ & $82.5$ & $61.8$ & $\mathbf{87.4}$ & $92.6$ & $1.45$ \\
 & 9 & 291 & 1 & $83.5$ & $92.8$ & $86.3$ & $\mathbf{94.8}$ & $97.9$ & $1.28$ \\
 & 10 & 174 & 1 & $90.2$ & $95.4$ & $94.3$ & $\mathbf{98.3}$ & $99.4$ & $1.14$ \\
 & 11+ & 209 & 1 & $94.3$ & $96.2$ & $98.6$ & $\mathbf{99.0}$ & $99.5$ & $1.07$ \\
 & all & 2,500 & 3 & $48.1$ & $68.8$ & $43.1$ & $\mathbf{75.5}$ & $82.1$ & $1.53$ \\
\bottomrule
\end{tabular}\vspace{2pt}
\caption{\textbf{N-Queens results by number of queens given.} Valid completions (\%). \textit{Queens} represents the number of queens given. \textit{Solutions} represents the median solution count per input. \textit{Distinct} represents the mean number of distinct valid completions among the 20 rollouts.}
\label{tab:nqueens-full}
\end{table}

\begin{figure}[h]
\centering
\includegraphics[width=\linewidth]{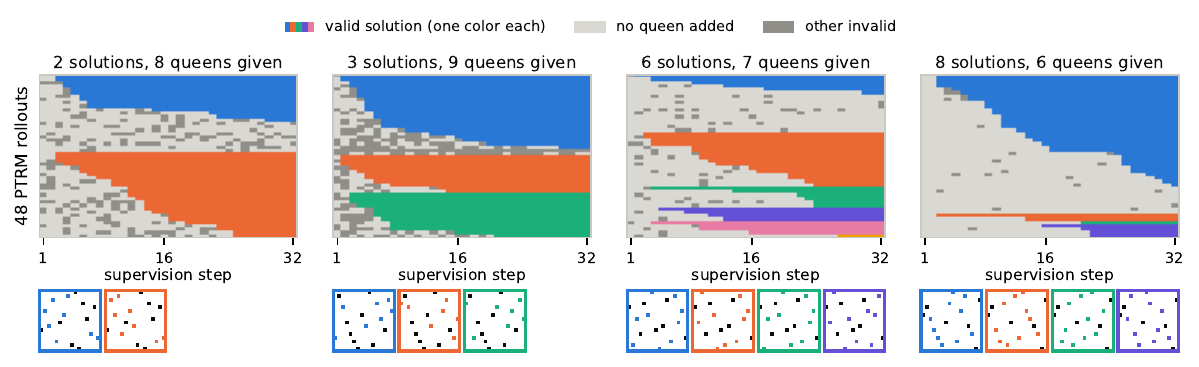}
\caption{\textbf{Rollout dynamics.} State of 48 PTRM rollouts ($\sigma{=}1.0$) at each supervision step on four N-Queens ($16{\times}16$) puzzles with 2 to 8 valid solutions (top), and some of the solutions reached, with given queens in black (bottom).}
\label{fig:nqueens-rollouts}
\end{figure}

Rollouts generally start by outputting the input board (with no queen added) and eventually commit to one solution. Different rollouts can commit to different solutions (Figure~\ref{fig:nqueens-rollouts}). This unfilled board is the most common invalid output, so it can sometimes get selected over valid solutions when using mode@$K$. Since the Q head scores those states near zero, best-Q@$K$ rarely selects it. Once valid, a rollout almost always stays fixed. On $500$ puzzles ($16{\times}16$ board), $98.2\%$ of the rollouts that reached a valid solution kept that exact solution through step 32 despite noise at every step. Rollouts commit later when the input has more solutions. The median first valid step goes from 3 on single-solution inputs to 8 on inputs with 4 queens removed (median of 118 solutions). Unlike deterministic TRM which also needs more steps (that must run sequentially) for multi-answer inputs, PTRM's rollouts run in parallel, so one of them is more likely to find the solution early. On the same 500 puzzles, best-Q over 16 rollouts reaches $58.0\%$ after 4 steps, which is above deterministic TRM's $50.8\%$ after 32 ($26.6\%$ after 4).

\subsection{Maze-Hard scored by path validity}
\label{app:maze-valid}
Maze-Hard stores a single shortest path per maze, but every test maze can have several, so simply matching the Maze-Hard labels can count some correct outputs as wrong. Table~\ref{tab:maze-valid} rescores the Maze-Hard runs of Sec.~\ref{sec:other-benchmarks} ($K{=}100$, $D{=}16$, $\sigma{=}1.0$, 3 seeds) on the 1000 test mazes, counting any valid shortest path as correct. Deterministic TRM goes from $83.8\%$ to $90.8\%$, best-Q@$K$ from $84.5\%$ to $91.3\%$ and pass@$K$ from $95.7\%$ to $98.0\%$. So the gap between pass@$K$ and best-Q@$K$ narrows from $11.2$ to $6.7$ points; about $40\%$ of it is explained by label matching. Maze-Hard remains the benchmark where Q-head selection is weakest, and mode@$K$ remains the better selection criterion there.

\begin{table}[h]
\centering
\small
\begin{tabular}{lcccc}
\toprule
Scoring & TRM & mode@$K$ & best-Q@$K$ & pass@$K$ \\
\midrule
label match & $83.8$ & $87.0$ & $84.5$ & $95.7$ \\
valid shortest path & $90.8$ & $93.2$ & $91.3$ & $98.0$ \\
\bottomrule
\end{tabular}\vspace{2pt}
\caption{\textbf{Maze-Hard by label match vs. path validity.} Results on the 1000 test mazes with $K{=}100$, $D{=}16$, and $\sigma{=}1.0$.}
\label{tab:maze-valid}
\end{table}

\section{Q-guided Langevin sampling}
\label{app:langevin}

We initially explored Langevin sampling (using the Q head gradient) as a more principled exploration mechanism than the Gaussian noise injection used in PTRM. The idea is to better guide the stochastic search by additionally steering each rollout (using the Q head gradient) toward regions of high Q value. We ultimately found that the gain from this approach was entirely attributable to the Langevin noise term, with the gradient component contributing nothing measurable on top of the equivalent recurrent noise of Sec.~\ref{sec:width_scaling}. We document the approach here as a negative result.

\begin{figure}[h]
\centering
\includegraphics[width=\linewidth]{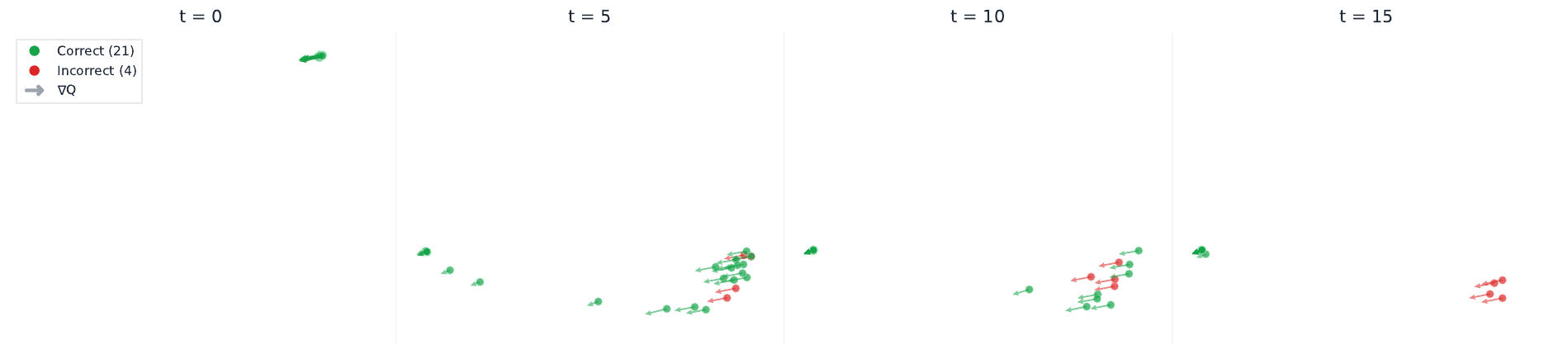}
\caption{$y$ latents and their $\nabla_z f_Q$ gradients projected into the
principal plane at several recursive/supervision steps, for multiple rollouts
(using recurrent noise) of a single puzzle (correct rollouts in green,
incorrect in red). Arrows are drawn at each latent in the direction of
$\nabla_z f_Q$. From both good-basin and bad-basin latents, gradients
point toward the good-basin region. This visualization motivated the
Langevin sampling experiment described below.}
\label{fig:gradient_snapshots}
\end{figure}

\paragraph{Motivation.}
The Q head is trained as a correctness predictor over latent states. Let
$f_Q(z)$ denote the head's scalar output. We treated $E(z) = -\log \mathrm{sigmoid}(f_Q(z))$ as an energy function over latent space. Empirical observations during early experiments suggested that regions of low $E$ correspond to good basins from which the decoded answer is likely correct. PCA visualizations of the latent dynamics showed that $\nabla_z f_Q$ points toward the good-basin region from both good-basin (correct) and bad-basin (incorrect) latents (Figure~\ref{fig:gradient_snapshots}). This made $\nabla_z f_Q$ look like a valuable direction along which to push latents.

\paragraph{Method.}
We sample from the target distribution
$p(z) \propto e^{-E(z)} = \mathrm{sigmoid}(f_Q(z))$ via Langevin
dynamics where at the end of each deep recursion step
$t = 1, \ldots, D$ we apply $N$ Langevin steps to the latent,
\[
z \leftarrow z - \eta\, \nabla_z E(z) + \sqrt{2\eta}\, \xi,
\quad \xi \sim \mathcal{N}(0, I),
\]
The number of Langevin steps $N$ is the additional scaling axis under this scheme.

\paragraph{Tractable gradient computation.}
TRM's original Q head is a linear projection on a single token,
$f_Q(y) = w^\top y[:, 0] + b$, so its gradient with respect to this
head's input is a constant vector independent of $z$. For
$\nabla_z f_Q$ to be input-dependent, the gradient must flow back
through the last latent recursion. This works but requires
backpropagating through a full latent recursion at every Langevin step,
which scales poorly with $N$. To make guidance tractable for large $N$,
we replaced the linear Q head with an attention-pooled variant that reads
the full latent and produces a scalar through a small nonlinear network.
With this head, $\nabla_z f_Q$ can be computed by backpropagating through
the head alone, which is $\sim$8$\times$ faster per step and does not sacrifice accuracy.

\paragraph{The gain came from the noise, not the gradient.}
Comparing Langevin sampling against a noise-only ablation (with the same
$\sqrt{2\eta}\, \xi$, but with the $-\eta\, \nabla_z E(z)$ term zeroed out) produced essentially identical accuracy at matched $N$. The gradient component contributed nothing measurable on top of the equivalent recurrent noise. This
prompted us to focus on the noise-only formulation in Sec.~\ref{sec:width_scaling}, which is much more impactful since it is: 1) significantly simpler (no retraining, no test-time backpropagation), 2) applicable to any TRM checkpoint out of the box, and 3) equally effective.


\clearpage
\newpage
\section*{NeurIPS Paper Checklist}

\begin{enumerate}

\item {\bf Claims}
    \item[] Question: Do the main claims made in the abstract and introduction accurately reflect the paper's contributions and scope?
    \item[] Answer: \answerYes{} 
    \item[] Justification: The abstract and introduction state PTRM's contributions and results, which are supported by the experiments in Sections 4–5 and the appendix. Limitations are discussed in Sec. \ref{sec:conclusion}.

\item {\bf Limitations}
    \item[] Question: Does the paper discuss the limitations of the work performed by the authors?
    \item[] Answer: \answerYes{} 
    \item[] Justification: Yes, we addressed this in Sec. \ref{sec:conclusion}.

\item {\bf Theory assumptions and proofs}
    \item[] Question: For each theoretical result, does the paper provide the full set of assumptions and a complete (and correct) proof?
    \item[] Answer: \answerNA{} 
    \item[] Justification: N/A

    \item {\bf Experimental result reproducibility}
    \item[] Question: Does the paper fully disclose all the information needed to reproduce the main experimental results of the paper to the extent that it affects the main claims and/or conclusions of the paper (regardless of whether the code and data are provided or not)?
    \item[] Answer: \answerYes{} 
    \item[] Justification: We have included all of the model details and experimental hyperparameter in Appendix~\ref{app:implementation-details}.

\item {\bf Open access to data and code}
    \item[] Question: Does the paper provide open access to the data and code, with sufficient instructions to faithfully reproduce the main experimental results, as described in supplemental material?
    \item[] Answer: \answerYes{} 
    \item[] Justification: The code will be published upon the acceptance.

\item {\bf Experimental setting/details}
    \item[] Question: Does the paper specify all the training and test details (e.g., data splits, hyperparameters, how they were chosen, type of optimizer) necessary to understand the results?
    \item[] Answer: \answerYes{} 
    \item[] Justification: We have included all of the model details and experimental hyperparameter in Appendix~\ref{app:implementation-details}.

\item {\bf Experiment statistical significance}
    \item[] Question: Does the paper report error bars suitably and correctly defined or other appropriate information about the statistical significance of the experiments?
    \item[] Answer: \answerYes{} 
    \item[] Justification: In this work, the PTRM results are reported across multiple seeds. We additionally report PPBench accuracy on a validation set, which is much larger than the golden set and provides tighter estimates.

\item {\bf Experiments compute resources}
    \item[] Question: For each experiment, does the paper provide sufficient information on the computer resources (type of compute workers, memory, time of execution) needed to reproduce the experiments?
    \item[] Answer: \answerYes{} 
    \item[] Justification: We have included all of the compute resources in Appendix~\ref{app:implementation-details}.
    
\item {\bf Code of ethics}
    \item[] Question: Does the research conducted in the paper conform, in every respect, with the NeurIPS Code of Ethics \url{https://neurips.cc/public/EthicsGuidelines}?
    \item[] Answer: \answerYes{} 
    \item[] Justification: The research adheres to ethics guidelines.

\item {\bf Broader impacts}
    \item[] Question: Does the paper discuss both potential positive societal impacts and negative societal impacts of the work performed?
    \item[] Answer: \answerNA{}
    \item[] Justification: This is foundational research on test-time compute for small models on reasoning puzzles, with no direct path to deployment, so we do not foresee direct positive or negative societal impacts.
    
\item {\bf Safeguards}
    \item[] Question: Does the paper describe safeguards that have been put in place for responsible release of data or models that have a high risk for misuse (e.g., pre-trained language models, image generators, or scraped datasets)?
    \item[] Answer: \answerNA{} 
    \item[] Justification: N/A

\item {\bf Licenses for existing assets}
    \item[] Question: Are the creators or original owners of assets (e.g., code, data, models), used in the paper, properly credited and are the license and terms of use explicitly mentioned and properly respected?
    \item[] Answer: \answerYes{} 
    \item[] Justification: Existing assets are properly cited.

\item {\bf New assets}
    \item[] Question: Are new assets introduced in the paper well documented and is the documentation provided alongside the assets?
    \item[] Answer: \answerNA{} 
    \item[] Justification: N/A

\item {\bf Crowdsourcing and research with human subjects}
    \item[] Question: For crowdsourcing experiments and research with human subjects, does the paper include the full text of instructions given to participants and screenshots, if applicable, as well as details about compensation (if any)? 
    \item[] Answer: \answerNA{} 
    \item[] Justification: N/A

\item {\bf Institutional review board (IRB) approvals or equivalent for research with human subjects}
    \item[] Question: Does the paper describe potential risks incurred by study participants, whether such risks were disclosed to the subjects, and whether Institutional Review Board (IRB) approvals (or an equivalent approval/review based on the requirements of your country or institution) were obtained?
    \item[] Answer: \answerNA{} 
    \item[] Justification: N/A

\item {\bf Declaration of LLM usage}
    \item[] Question: Does the paper describe the usage of LLMs if it is an important, original, or non-standard component of the core methods in this research? Note that if the LLM is used only for writing, editing, or formatting purposes and does \emph{not} impact the core methodology, scientific rigor, or originality of the research, declaration is not required.
    \item[] Answer: \answerNA{} 
    \item[] Justification: N/A

\end{enumerate}

\end{document}